\documentclass[10pt, a4paper]{article}

\usepackage{lrec-coling2024} 

\usepackage{color}
\usepackage{xcolor}
\usepackage{booktabs}
\usepackage{longtable}
\usepackage{multirow}
\usepackage[hang,flushmargin]{footmisc}
\usepackage{dblfloatfix}

\usepackage{linguex}
\usepackage{dingbat}
\usepackage{amsmath}
\usepackage{cleveref} 

\newcommand{\dri}[1]{\textcolor{red}{\bf\small }}
\newcommand{\ms}[1]{\textcolor{red}{\bf\small }}
\newcommand{\ees}[1]{\textcolor{red}{\bf\small }}
\newcommand{\mt}[1]{\textcolor{red}{\bf\small }}
\newcommand{\mn}[1]{\textcolor{red}{\bf\small }}
\newcommand{\peter}[1]{\textcolor{red}{\bf\small }}
\newcommand{\ks}[1]{\textcolor{red}{\bf\small }}

\title{Evaluating the IWSLT2023 Speech Translation Tasks: \\ Human Annotations, Automatic Metrics, and Segmentation}

\name{Matthias Sperber$^{1}$, Ondřej Bojar$^2$, Barry Haddow$^3$, Dávid Javorský,$^2$  \\ {\bf \large  Xutai Ma$^4$, Matteo Negri$^5$, Jan Niehues$^6$, Peter Polák$^2$}\\ {\bf \large Elizabeth Salesky$^{1,7}$, Katsuhito Sudoh$^8$, Marco Turchi$^9$}} 

\address{$^1$Apple, $^2$Charles~U., $^3$U.~Edinburgh, $^4$Meta, $^5$FBK, $^6$KIT, $^7$JHU, $^8$NAIST, $^9$Zoom \\
         sperber@apple.com, bojar@ufal.mff.cuni.cz, bhaddow@ed.ac.uk, javorsky@ufal.mff.cuni.cz, \\
         xutaima@meta.com, negri@fbk.eu, jan.niehues@kit.edu, polak@ufal.mff.cuni.cz \\
         esalesky@jhu.edu, sudoh@is.naist.jp, marco.turchi@zoom.us \\}

\abstract{
Human evaluation is a critical component in machine translation system development and has received much attention in text translation research.
However, little prior work exists on the topic of human evaluation for speech translation, which adds additional challenges such as noisy data and segmentation mismatches.
We take first steps to fill this gap by conducting a comprehensive human evaluation of the results of several shared tasks from the last International Workshop on Spoken Language Translation (IWSLT 2023).
We propose an effective evaluation strategy based on automatic resegmentation and direct assessment with segment context.
Our analysis revealed that: \textit{1)} the proposed evaluation strategy is robust and scores well-correlated with other types of human judgements; \textit{2)} automatic metrics are usually, but not always, well-correlated with direct assessment scores; and \textit{3)} \textsc{COMET} as a slightly stronger automatic metric than \textsc{chrF}, despite the segmentation noise introduced by the resegmentation step systems. 
We release the collected human-annotated data in order to encourage further investigation.
 \\ \newline  }

\begin{document}

\maketitleabstract

\section{Introduction}

Human evaluation plays a critical role in the development of translation systems and, although costly, it serves as a gold standard against which to calibrate automatic metrics.\footnote{This is especially true for \textit{trainable} metrics, which have recently started to outperform traditional metrics \cite{freitag-etal-2022-results}.} Moreover, it is important when in doubt about whether certain types of system behaviors are being accurately captured by automatic metrics. This is particularly relevant when the output quality is high, and automatic metrics become less reliable. For the case of machine translation for \textit{text} inputs (MT), a large body of prior research has investigated and improved procedures for manual evaluation, for example in the Conference on Machine Translation (WMT) series of shared tasks \citep{kocmi-etal-2022-findings}. Many of the popular MT metrics have been shown to have a high correlation with human evaluation results \cite{freitag-etal-2022-results}.

In this paper, we focus on evaluation in \textit{speech} translation (ST), which introduces additional complicating factors, such as erroneous automatic segmentation, dealing with noisy inputs, conversational and disfluent language, and special downstream requirements, such as simultaneous translation, and subtitling/dubbing constraints. To our knowledge, little prior work has focused on meta-analysis of human evaluation for speech translation. Consequently, it remains unclear to what degree procedures for human and automatic evaluation from MT can be transferred to the ST scenario.

As a first step toward developing trusted human evaluation procedures for ST, and following the successful related efforts at WMT, we conducted a manual evaluation of several shared tasks from the International Workshop for Spoken Language Translation (IWSLT) 2023. Tasks include translation of presentations at the Conference of the Association of Computational Linguistics (ACL) and TED talks\footnote{Talks on technology, entertainment, and design, available at \url{www.ted.com}.} in several language pairs in offline, multilingual, and simultaneous translation conditions. In the offline and multilingual conditions, inputs are given as unsegmented long-form speech, requiring systems to apply automatic segmentation prior to translation. 

The shared tasks used test sets that partially overlap between different task conditions, providing the opportunity for analysis across various conditions.

For conducting our human evaluation, we choose an approach that allows the comparison of systems despite potentially mismatching segmentation by re-segmenting system outputs to a common segmentation and using segment context. To minimize costs, a random subset of segments is chosen for evaluation.

Through extensive analysis, we shed light on the current state of ST evaluation protocols contributing in three ways.
First, we confirm that our collected direct assessment (DA) scores are well-correlated with additionally collected annotations, namely Multidimensional Quality Metric (MQM) and continuous ratings, indicating that the proposed evaluation procedure is sound and reliable to be used in future work. 
Second, we show that the correlation between human evaluation and automatic metrics is often, but not always, high. We conclude that these metrics can generally be trusted, but that human evaluation should be continually run side-by-side for further verification. 

Third, we show that COMET \cite{rei-etal-2020-comet} has higher correlation than other automatic metrics despite potential robustness issues due to noisy automatic segmentation. This shows promise in using trainable metrics like COMET for speech translation scenarios, although more detailed follow-up experiments is needed. 

To the best of our knowledge, the released annotations would be the first publicly available annotations of their kind for speech rather than text translation, facilitating the comparison of evaluation methodology between modalities.\footnote{Data is available in WMT format under \url{https://huggingface.co/datasets/IWSLT/da2023}.}

\section{Task Description}

\begin{table}[tb]
\centering
\begin{tabular}{lccccr}
\toprule
Task & Offline    & Multilingual & Simultaneous \\
\midrule
TED  & \checkmark &              & \checkmark   \\
ACL  & \checkmark & \checkmark   &              \\
     
\bottomrule
\end{tabular}
\caption{Domains used by the 3 shared tasks.}
\label{tab:tasks_domains}
\end{table}

In this section, we provide a bird's eye view of the IWSLT 2023 tasks considered in our study. Accordingly, we concentrate on the \textit{task conditions}, the \textit{data} used, and the \textit{automatic evaluation} protocols.
\subsection{Conditions}

\subsubsection{Offline}

The Offline Speech Translation Task aimed to explore automatic methods for translating spoken language in one language into written text in another language. This could be achieved through either cascaded solutions involving pipelined automatic speech recognition and machine translation (MT) systems as core components, or end-to-end approaches that directly translate the audio bypassing the intermediate transcription step. Recent results \cite{bentivogli-etal-2021-cascade} have shown that the performance of end-to-end models is becoming comparable to that of cascade solutions, but the best-performing technology has yet to be identified. 

In the 2023 edition, the Offline Speech Translation Task not only addressed the question of whether the cascade solution remains the prevailing technology but it also assessed the systems submitted in more intricate and demanding situations. These included multiple speakers, non-native speakers, different accents, varying recording quality, specialized terminology, controlled interaction with a second speaker, and spontaneous speech.

Submitted systems used a variety of approaches, experimenting with both constrained and unconstrained training data conditions as well as use of pretrained large language models.

\subsubsection{Multilingual}

The Multilingual Task focused particularly on the capability to translate to a wide variety of target languages in a realistic use case: with long-form audio requiring segmentation, diverse speaker accents, and domain-specific terminology. 
The task used the full set of ten target languages from the ACL 60-60 evaluation sets,\footnote{Arabic, Mandarin Chinese, Dutch, French, German, Japanese, Farsi, Portuguese, Russian, and Turkish.} of which three are considered in this paper (see Section~\ref{sec:test_data}).
Teams were required to submit to all language pairs, though not restricted to multilingual modelling approaches, such that all approaches could be compared across all ten target languages. 
There were comparisons between both cascaded and end-to-end systems, a variety of pretrained models, as well as multilingual and single language pair finetuning.

\subsubsection{Simultaneous}

Simultaneous translation, also known as real-time or online translation,
is the task of generating translations incrementally given partial input only. 
It enables low-latency applications such as simultaneous interpretation in personal traveling and international conferences. 
In the 2023 edition, a submission qualified as a simultaneous system if the latency is no greater than average lagging \cite{ma-etal-2019-stacl} of 2 seconds.

The organizers of the shared task proposed two tracks, speech-to-text and speech-to-speech, over three language pairs, 
English to German, Chinese and Japanese.
The human evaluation is only conducted over speech-to-text systems.

\subsection{Test Data}
\label{sec:test_data}
As shown in Table~\ref{tab:tasks_domains}, test data came from the TED domain (for offline and simultaneous tasks), and ACL domain (for offline and multilingual tasks), as detailed below. For the offline and multilingual task, this data was given as long-form speech, requiring systems to use automatic segmentation strategies. The simultaneous task provided input data according to the reference segmentation. For both datasets, our analysis concentrates on German, Japanese, and Mandarin Chinese as target language, and English as source language.

\begin{figure*}[t]
\includegraphics[width=\textwidth]{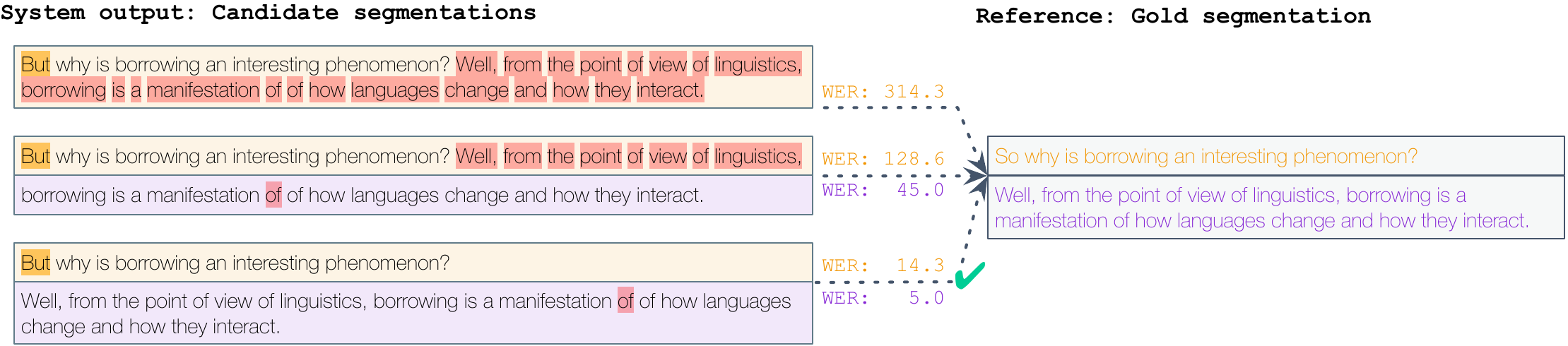}
\caption{To parallelize system outputs with  reference translations, we re-segment hypotheses by minimizing the overall WER to the reference with \textsc{mwerSegmenter}, which supports many-to-many realignment.}
\label{fig:resegmentation-mwer}
\end{figure*}

\subsubsection{TED}
\label{sec:ted_data}

For the TED scenario, the test sets are built starting from 42 talks that are not yet part of the en-de section of the current public release of the TED-derived MuST-C corpus \cite{CATTONI2021101155}. 
From this material, talks that have been translated into Japanese and Chinese were selected to build the en-ja and en-zh test sets, which consist of 37 and 38 talks respectively.
 
There are two different types of target-language references used in the evaluation campaign. The first type is the original TED translations, which come in the form of subtitles. Because of TED's subtitling guidelines,\footnote{\url{https://www.ted.com/participate/translate/guidelines}} these translations may contain compressed or omitted content, making them less literal compared to unconstrained translations. The second type is unconstrained 
translations, which were created from scratch by professionals, following typical translation guidelines. These translations are therefore exact, meaning they are literal and have proper punctuation.

\subsubsection{ACL}

The second test data set is based on the ACL 60-60 evaluation sets released by \citet{salesky-etal-2023-evaluating}, which  
are composed of technical presentations from ACL 2022 given in English by diverse speakers across different paper topics. 
The talks were manually sentence segmented, transcribed, and translated into ten target languages from the 60/60 initiative.\footnote{\url{www.2022.aclweb.org/dispecialinitiative}}

The evaluation sets contain 1 hour of one-to-many parallel speech, transcripts, and translations per language pair.

The ACL test sets introduce additional challenges beyond those in the traditional TED setting, including the presence of non-native speakers with diverse accents, varying recording quality, and the strong presence of technical terminology.

\section{Evaluation}

\subsection{Resegmentation}
\label{sec:resegmentation}

To follow realistic use conditions, no reference segmentation was provided for the offline and multilingual tasks; rather, participants were required to process each talk as a whole. 
To this end, participants used different tools (either publicly available or proprietary) to automatically segment the audio for downstream processing. 

Consequently, the segmentation in each submitted system can differ significantly from the reference segmentation. 
To parallelize outputs and references for evaluation, we re-segmented translation hypotheses following \citet{matusov-etal-2005-evaluating} 
by exploiting WER alignment to the reference translation with the tool \textsc{mwerSegmenter}.\footnote{\url{https://www-i6.informatik.rwth-aachen.de/web/Software/mwerSegmenter.tar.gz}} 
In this step, the hypotheses are monotonically re-segmented in order to minimize the global WER to the reference translation, with candidate boundaries determined by tokenization; in our case, word-level for German and character-level for Chinese and Japanese. 
An illustrative example from the ACL data is shown in \autoref{fig:resegmentation-mwer}. 
All subsequent evaluation, both automatic and human, used the re-segmented hypotheses which are now parallel to the references.

\subsection{Automatic Metrics}
\label{sec:task_auto_eval}

For automatic evaluation, the evaluation campaign focused on standard lexical and model-based machine translation evaluation metrics, namely BLEU \cite{papineni-etal-2002-bleu}, \textsc{chrF} \cite{popovic-2015-chrf} and COMET\footnote{Unbabel/wmt22-comet-da} \cite{rei-etal-2020-comet}. For conciseness, here we focus on the last two, as BLEU has been shown to be less reliable than \textsc{chrF} \cite{freitag-etal-2022-results}. As is common practice in speech translation evaluation, automatic metrics are computed using the re-segmented hypotheses as described in Section~\ref{sec:resegmentation}.

\begin{figure*}[t]
\begin{small}
\begin{tabular}{ll}

\small{Source text:}                & \small{\textbf{We assume the precision of quantities are known.}}                                                               \\
\hline
\small{Reference target text:}      & 
    \begin{tabular}{@{}c@{}c@{}c@{}c@{}c@{}c@{}c@{}c@{}c@{}c@{}c@{}}
    \textit{We\hspace{0.15cm}} & \textit{go}                   & \multicolumn{2}{l}{\textit{under assumption,}}                                     & \textit{that}        & \textit{the}       & \textit{precision}  & \textit{of the} & \textit{sets}         & \textit{known}         & \textit{is.}  \\
    \textbf{Wir}               & \textbf{gehen}\hspace{0.15cm} & \textbf{davon}\hspace{0.15cm} & \textbf{aus}\hspace{0.15cm} & \textbf{dass\hspace{0.15cm}} & \textbf{die}\hspace{0.15cm} & \textbf{Genauigkeit}         & \textbf{der}             & \textbf{Mengen}\hspace{0.15cm} & \textbf{bekannt}\hspace{0.15cm} & \textbf{ist.}  \\
    \end{tabular} 
    \\
\hline
\small{Reseg.\ sys.\ output:} & 
    \begin{tabular}{@{}c@{}c@{}c@{}c@{}c@{}c@{}c@{}c@{}c@{}c@{}}
    \textit{we\hspace{0.15cm}} & \multicolumn{2}{c}{\textit{under assumption,}}            & \textit{that}        & \textit{the}       & \textit{precision}  & \textit{of the} & \textit{sets}         & \textit{known}         & \textit{is,}  \\
    \textbf{\textcolor{red}wir}               & \multicolumn{2}{c}{\textcolor{orange}{\textbf{davon aus,}}}\hspace{0.15cm} &  \textbf{dass\hspace{0.15cm}} & \textbf{die}\hspace{0.15cm} & \textbf{Präzision}         & \textbf{von}             & \textbf{Mengen}\hspace{0.15cm} & \textbf{bekannt}\hspace{0.15cm} & \textbf{ist\textcolor{purple},}  \\
    \end{tabular} 
    \\

\hline
\small{Previous sys. output:}     & 
    \begin{tabular}{@{}c@{}c@{}c@{}c@{}c@{}c@{}c@{}c@{}c@{}c@{}c@{}c@{}}
    \textit{It\hspace{0.15cm}} & \textit{holds}            & \textit{thus}        & \textit{also}       & \textit{certain}  & \textit{assumptions.} & \textit{As}         & \textit{in}         & \textit{prior} & \textit{works} & \textit{go}  \\
    \textbf{Es}\hspace{0.15cm}               & \textbf{gelten}\hspace{0.15cm} &  \textbf{also\hspace{0.15cm}} & \textbf{auch}\hspace{0.15cm} & \textbf{bestimmte}         & \textbf{Annahmen.}             & \textbf{Wie}\hspace{0.15cm} & \textbf{in}\hspace{0.15cm} & \textbf{früheren} \hspace{0.15cm}& \textbf{Arbeiten}\hspace{0.15cm} & \textcolor{orange}{\textbf{gehen}} &   \\
    \end{tabular} 
    \\

\hline
\small{Next system output:}         & 

    \begin{tabular}{@{}c@{}c@{}c@{}c@{}c@{}c@{}c@{}c@{}c@{}c@{}c@{}c@{}}
    \textit{and\hspace{0.15cm}} & \textit{consider}            & \textit{only}        & \textit{basic}       & \textit{operators}  & \textit{like} & \textit{additon,}         & \textit{subtraction,}         & \textit{$\cdots$}   \\
    \textbf{\textcolor{red}und}\hspace{0.15cm}               & \textbf{betrachten}\hspace{0.15cm} &  \textbf{nur\hspace{0.15cm}} & \textbf{grundlegende}\hspace{0.15cm} & \textbf{Operatoren}\hspace{0.15cm}         & \textbf{wie}\hspace{0.15cm}             & \textbf{Addition,}\hspace{0.15cm} & \textbf{Subtraktion,}\hspace{0.15cm} & \textbf{$\cdots$} \hspace{0.15cm} \\
    \end{tabular} 
    \\

\end{tabular}
\caption{Example for annotation of re-segmented English-German system outputs. Gloss is provided in italics. The human annotator, if looking only at the re-segmented system output of the current sentence, would find the \textcolor{orange}{main verb missing}, the beginning of the sentence \textcolor{red}{not capitalized}, and the sentences \textcolor{purple}{ending in a comma instead of full stop}, and penalize the sentence accordingly. As a remedy, we also present system outputs for adjacent segments to the annotator. The main verb (\textcolor{orange}{\textit{gehen}}) is located in the system output for the previous sentence, and by being presented with the context, the annotator will be able to see that this translation has no grammatical issues nor problems with casing or punctuation.}
\label{tab:resegmentation-annotation}
\end{small}
\end{figure*}

\subsection{Human Evaluation}

Conducting human evaluation is important for a number of reasons. Among others, prior work \cite{barrault-etal-2019-findings} has observed that for highly accurate translation systems, human evaluators often rank systems differently than automatic metrics, indicating that automatic metrics alone may not be reliable enough in many situations. Moreover, recent speech translation research places much focus on comparing two different modeling paradigms, i.e.\ direct vs.\ cascaded ST. This is reminiscent of an earlier situation in MT research, where automatic metrics, when used to compare different 
MT paradigms (rule-based, statistical, neural), were found to be less reliable than for comparing similar systems 
built under the same paradigm \cite{bojar-etal-2016-findings}.  Similarly, it could be conceivable that automatic metrics may suffer from systematic biases that would make comparison across ST modeling paradigms difficult. However, whether (and to what degree) these modeling paradigms lead to systematically different translations is currently not well understood.\footnote{Some initial work exists, for instance \newcite{bentivogli-etal-2021-cascade} do not find significant differences in preference, while \newcite{gaido-etal-2020-breeding} investigate speaker gender awareness as a concept that only direct models can learn in theory, since cascaded models lose this information through reliance on the intermediate transcription.\looseness=-1}
In light of this,
we argue that more emphasis is needed on human evaluation for comparing ST paradigms.
Beyond such considerations, setting up a gold procedure 
is important for a variety of reasons, including establishing or training better automatic metrics.

\subsubsection{Direct assessment}
\label{sec:da}

We conduct a source-based DA \cite{graham-etal-2013-continuous,iwslt:2017,akhbardeh-etal-2021-findings}, where annotators are shown both the source text and the translated target text.

For DA, we again make use of the automatic re-segmentation as outlined in 
\S\ref{sec:resegmentation}, which has several important benefits. First, it makes possible to choose only a subset of segments for manual annotation, which is often desirable for cost considerations while ensuring that the segmentation and choice of segments remain comparable across the evaluated systems. Second, it allows computing segment-level correlation with automatic metrics, thanks to consistent segmentation. Finally, it frees us from having to apply weighted averaging schemes when aggregating segment scores, in order to make up for the possibility of different systems using different segment granularity.

Figure~\ref{tab:resegmentation-annotation} shows an example of how DA based on resegmented system outputs looks in practice. 

Note that the re-segmentation step described above may lead to cases where annotators are presented with segments of automatic translations that start or stop in the middle of a sentence (see example in Figure~\ref{tab:resegmentation-annotation}). 
Such apparent problems in the translation output may originate simply 
from the different segmentation but may be perfectly valid when considered in context. To avoid penalizing such situations, we provided translators not only with the source sentence and system translation but also with the system translation of the previous and following segments. The precise annotation instructions are given 
in the appendix. Providing more context to human annotators is additionally motivated by prior research  demonstrating higher annotation quality by showing document context to annotators \cite{grundkiewicz-etal-2021-user}.

Assessments were performed on a continuous scale between 0 and 100. Neither video nor audio context was provided.
Segments were shuffled and randomly assigned to annotators to avoid bias related to the presentation order.
Annotations were conducted through a trusted vendor by a total of 77 professional translators (33 for English-German, 11 for English-Japanese, 33 for English-Chinese). Annotators were fluent in the source language and native in the target language. 

DA scores were collected for all three considered language pairs, i.e.\ English-German, English-Japanese, and English-Chinese. For the TED domain, DA scores are collected over a subset of 1,000 randomly chosen segments. The same segments are chosen for  all the systems.

\begin{table*}[tb]
\centering
\begin{tabular}{lccccr}
\toprule
        Task & Language & Domain & Systems &   Segments  & Tokens \\
\midrule
     Offline & en-de    & TED &        $7$ &      $7000$ & $115633$\\
Multilingual & en-de    & ACL &        $3$ &      $1248$ & $22128$\\
     Offline & en-de    & ACL &        $7$ &      $2912$ & $51632$\\
Simultaneous & en-de    & TED &        $5$ &      $5000$ & $82595$\\
Simultaneous & en-ja    & TED &        $4$ &      $4000$ & $66076$\\
     Offline & en-ja    & TED &        $5$ &      $5000$ & $82595$\\
     Offline & en-ja    & ACL &        $5$ &      $2080$ & $36880$\\
Multilingual & en-ja    & ACL &        $3$ &      $1248$ & $22128$\\
     Offline & en-zh    & TED &        $8$ &      $8000$ & $132152$\\
     Offline & en-zh    & ACL &        $8$ &      $3328$ & $59008$\\
Multilingual & en-zh    & ACL &        $3$ &      $1248$ & $22128$\\
Simultaneous & en-zh    & TED &        $2$ &      $2000$ & $33038$\\

\bottomrule
\end{tabular}
\caption{Data statistics for collected direct assessments: Number of systems, segments, and source side (reference transcript) tokens for each combination of task, language pair, and domain.}
\label{tab:da_datasize}
\end{table*}

\begin{table*}[tb]
\centering
\begin{tabular}{lcccrlrl}
\toprule
    Task     &Language&Domain&Systems&\multicolumn{2}{c}{$\rho$}  & \multicolumn{2}{c}{$r$} \\
\midrule

Offline      & en-de & TED & $7$ & $\mathbf{0.99}$ & $(p{=}0.00)$ &         $0.71$  & $(p{=}0.07)$ \\
Multilingual & en-de & ACL & $3$ &         $0.98$  & $(p{=}0.12)$ & $\mathbf{1.00}$ & $(p{=}0.00)$ \\
Offline      & en-de & ACL & $7$ & $\mathbf{0.94}$ & $(p{=}0.00)$ &         $0.75$  & $(p{=}0.05)$ \\
Simultaneous & en-de & TED & $5$ &         $0.75$  & $(p{=}0.15)$ &         $0.50$  & $(p{=}0.39)$ \\
Simultaneous & en-ja & TED & $4$ & $\mathbf{0.99}$ & $(p{=}0.01)$ &         $0.20$  & $(p{=}0.80)$ \\
Offline      & en-ja & TED & $5$ & $\mathbf{0.99}$ & $(p{=}0.00)$ & $\mathbf{0.90}$ & $(p{=}0.04)$ \\
Offline      & en-ja & ACL & $5$ & $\mathbf{0.99}$ & $(p{=}0.00)$ &         $0.60$  & $(p{=}0.28)$ \\
Multilingual & en-ja & ACL & $3$ &         $0.97$  & $(p{=}0.16)$ &         $0.50$  & $(p{=}0.67)$ \\
Offline      & en-zh & TED & $8$ & $\mathbf{0.96}$ & $(p{=}0.00)$ & $\mathbf{0.79}$ & $(p{=}0.02)$ \\
Offline      & en-zh & ACL & $8$ & $\mathbf{0.98}$ & $(p{=}0.00)$ & $\mathbf{0.98}$ & $(p{=}0.00)$ \\
Multilingual & en-zh & ACL & $3$ & $\mathbf{1.00}$ & $(p{=}0.03)$ & $\mathbf{1.00}$ & $(p{=}0.00)$ \\
Simultaneous & en-zh & TED & $2$ & $\mathbf{1.00}$ & $(p{=}1.00)$ &         $1.00$  & $(p{=}\text{n/a})$ \\
\bottomrule
\end{tabular}
\caption{Pearson correlation ($\rho$) and Spearman correlation ($r$) of DA scores vs.\ \textsc{chrF} scores.}
\label{tab:da_sys_corr}
\end{table*}

\subsubsection{MQM}

\label{sec:mqm}
As part of the evaluation campaign, a portion of the submissions were annotated using Multidimensional Quality Metrics \cite[MQM;][]{lommel_2014}. 

MQM has been used in the WMT shared task series in recent years \cite{freitag-etal-2021-results} and is promising for detailed analyses of translation results. 
We include this data in our analysis in order to verify the quality of the collected DA scores.
The MQM evaluation was focused only on English-to-Japanese simultaneous translation. 
See \citet{agrawal-etal-2023-findings} for more details.

\subsubsection{Continuous Rating}
\label{sec:countinuous_rating}

In addition, the IWSLT 2023 evaluation campaign also included 

a human evaluation study using the \emph{Continuous Rating} method \cite{javorsky-etal-2022-continuous} for the English-to-German simultaneous translation within the TED domain. 
This evaluation method was selected as it has been specifically designed for the simultaneous translation regime. Again, we include this data in our analysis in order to verify the quality of the collected DA scores. 

CR replicates the real-world translation scenario, in which the speaker receives incomplete translations while delivering the speech. Native German speakers fluent in English were assigned to listen to the source audio in English and continuously evaluate the quality of the German translation. 
The translation grew incrementally with respect to the timestamps corresponding to the source audio as the system generated each word. All systems followed the same reference segmentation of the source audio, but the human evaluation was on the talk level. For more details about the evaluation process, please refer to \citet{agrawal-etal-2023-findings} and \citet{anastasopoulos-etal-2022-findings}.

\section{Main Results}
\begin{table*}[tb]
\centering
\begin{tabular}{lcccccccc}
\toprule
  & & & \multicolumn{2}{c}{}  & \multicolumn{2}{c}{} & \multicolumn{2}{c}{\small Original TED refs.} \\
 \small Task & \small Lang.  & \small Dom.  &  $\rho_{\small_{\textit{DA}/\textsc{chrF}}}$ & $r_{\small_{\textit{DA}/\textsc{chrF}}}$ & $\rho_{\small_\textit{COMET/DA}}$ & $r_{\small_\textit{COMET/DA}}$ & $\rho_{\small_{\textit{DA}/\textsc{chrF}}}$ & $r_{\small_{\textit{DA}/\textsc{chrF}}}$ \\[5pt]
\midrule

\small Offline & \small en-de & \small TED & $\mathbf{0.99}$ &          $0.71$ & $\mathbf{0.99}$ &         $0.68$  & $\mathbf{0.97}$ & $0.61$ \\
\small Multi   & \small en-de & \small ACL & $0.98$          & $\mathbf{1.00}$ & $\mathbf{1.00}$ & $\mathbf{1.00}$ &  --               & -- \\
\small Offline & \small en-de & \small ACL & $\mathbf{0.94}$ & $\mathbf{0.75}$ & $\mathbf{0.99}$ & $\mathbf{0.89}$ &  --               & -- \\
\small Simul   & \small en-de & \small TED & $0.93$          &         $0.60$  & $\mathbf{0.96}$ &         $0.80$  & $\mathbf{0.97} $ & $0.80$ \\
\small Simul   & \small en-ja & \small TED & $\mathbf{0.99}$ &         $0.20$  & $\mathbf{1.00}$ & $\mathbf{1.00}$  & $\mathbf{0.99}$ & $0.20$ \\
\small Offline & \small en-ja & \small TED & $\mathbf{0.99}$ & $\mathbf{0.90}$ & $\mathbf{1.00}$ &          $0.70$ & $\mathbf{0.99}$ & $\mathbf{0.90}$ \\
\small Offline & \small en-ja & \small ACL & $\mathbf{0.99}$ &  $0.60$         & $\mathbf{1.00}$ &          $0.70$ & -- & --  \\
\small Multi   & \small en-ja & \small ACL & $0.97$          &  $0.50$         & $\mathbf{1.00}$ & $\mathbf{1.00}$ & --          &  -- \\
\small Offline & \small en-zh & \small TED & $\mathbf{0.97}$ & $\mathbf{0.89}$ & $\mathbf{1.00}$ & $\mathbf{0.96}$ & $\mathbf{0.98}$ & $\mathbf{0.89}$ \\
\small Offline & \small en-zh & \small ACL & $\mathbf{0.98}$ & $\mathbf{0.96}$ & $\mathbf{1.00}$ & $\mathbf{1.00}$ & -- & -- \\
\small Multi   & \small en-zh & \small ACL & $\mathbf{1.00}$ & $\mathbf{1.00}$ &         $0.99$  & $\mathbf{1.00}$ & -- & -- \\
\small Simul   & \small en-zh & \small TED &         $1.00$  &         $1.00$  &         $1.00$  &         $1.00$  & $1.00$ &  $1.00$ \\

\bottomrule
\end{tabular}
\caption{Comparison of COMET and \textsc{chrF} correlation with DA scores, respectively. In addition, the DA/\textsc{chrF} correlation based on the original TED references are given.}
\label{tab:da_sys_corr_with_comet}
\end{table*}

\begin{table*}[tb]
    \centering
    \begin{tabular}{lllclll}
        \toprule
        \small Task & \small  Language & \small  Domain & \small Systems & $\rho_{\small_\textit{DA/MQM}}$ & $\rho_{\small_{\textsc{chrF}/\textit{MQM}}}$ & $\rho_{\small_{\textsc{COMET}/\textit{MQM}}}$\\[5pt]
        \midrule
        Simultaneous & en-ja & TED       & $4$                  & $-0.97\ (p{=}0.03)$
                     & $-0.99\ (p{=}0.01)$                  & $-0.98\ (p{=}0.02)$ \\
        \bottomrule
    \end{tabular}
    \caption{System-level correlation of DA, \textsc{chrF}, and COMET against MQM for the 107 segments subset.}
    \label{tab:mqm_en_ja}

\end{table*}

\begin{table*}[tb]
    \centering
    \begin{tabular}{lllclll}
        \toprule
        Task & Language & Domain & Systems & $\rho_\textit{DA/CR}$ & $\rho_{\textsc{chrF}/\textit{CR}}$ & $\rho_\textit{COMET/CR}$\\
        \midrule
        Simultaneous & en-de & TED       & $5$                  & $0.95\ (p{=}0.01)$ & $0.75\ (p{=}0.15)$ & $0.94\ (p{=}0.02)$ \\
        \bottomrule
    \end{tabular}
    \caption{System-level correlation of DA, \textsc{chrF}, and COMET against Continuous Rating for English-to-German simultaneous translation data.}
    \label{tab:cr_en_de}

\end{table*}

In our meta-evaluation, we compare human evaluation results with those computed with automatic metrics.
For the latter, we use the full evaluation set (416 sentences for the ACL domain, and around 2,000 for the TED domain depending on language pair), while DA on TED was restricted to 1,000 segments. Following \newcite{agrawal-etal-2023-findings}, we used the new, more natural TED references unless otherwise noted.

We rely on computing correlations as our main analysis tool. One challenge is that the number of data points (systems) per evaluated condition was between only 2 and 20, which on the lower end was sometimes too small to obtain statistically significant results. For a more robust analysis, we therefore present both Pearson linear correlation (denoted $\rho$) and Spearman rank correlation (denoted $r$). Intuitively, $\rho$ captures a metric's informativeness regarding the magnitude by which systems differ, while $r$ measures its reliability for ranking of systems, both of which are generally of interest. All correlations are based on system-level scores. Statistically significant correlations at $p{\leq}.05$ are shown in bold font throughout.

An overview of the systems under evaluation is provided by \newcite{agrawal-etal-2023-findings}. Detailed data size statistics are in Table~\ref{tab:da_datasize}.

\subsection{DA and Automatic Metrics}
\label{sec:results-da-automatic}

Table~\ref{tab:da_sys_corr} shows the correlation between \textsc{chrF} and DA scores. We observe generally high $\rho$, while $r$ tends to be lower in some cases. Note that no coefficients lower than 0.75 were statistically significant, suggesting an insufficient number of data points as the main reason for those very low correlation coefficients. The fact that results in many cases approach perfect correlation can be explained by observing that there was often a significant accuracy gap between the submitted systems \cite{agrawal-etal-2023-findings}. We speculate as another contributing factor that speech translation quality has not yet reached the level where automatic metrics become increasingly unreliable. This is in contrast to machine translation, where it is often found that for top-performing systems, automatic metrics have a poor correlation with human-produced assessments \citep{mathur-etal-2020-tangled}.

Table~\ref{tab:da_sys_corr_with_comet} shows, among others, the correlation between COMET and DA scores.
As can be observed in columns 4--7, COMET scores yielded higher correlations than \textsc{chrF} for all cases where both sides had statistically significant correlation coefficients. This indicates that COMET is robust enough to be applied to the speech translation scenario which includes additional challenges such as noisy sentence segmentation. However, prior work \cite[Appendix D1]{amrhein-haddow-2022-dont} casts some doubt on this conclusion, and a more thorough study of this issue is needed. Also, note that while the results show higher correlation for COMET scores with DA than \textsc{chrF} with DA, the magnitude of this improvement remains somewhat unclear, because \textsc{chrF} correlations are generally already so high that there is not much room for improvement. Further experiments are needed to better understand whether the extent to which trainable metrics such as COMET outperform string comparison based metrics such as \textsc{chrF} is comparable to the large gap observed in text translation \cite{freitag-etal-2022-results}.

To summarize: \textbf{DA is highly correlated with both automatic metrics in most but not all cases for our data, and COMET is found to outperform \textsc{chrF} in the speech translation scenario.}

\subsection{MQM}

The previous subsection investigates correlations between automatic metrics and human judgements in the form of DA scores, but we also wish to assess the soundness of the DA scores as a gold standard. While high correlation between DA and automatic metrics provides some positive indication, we now turn to MQM scores, available for portions of the English-Japanese data, as a more reliable means of verification of soundness of the DA scores. Table~\ref{tab:mqm_en_ja} shows correlations between the MQM score and other scores (DA, chrF, and COMET) at the system level. We observed clearly negative correlations\footnote{Note that an MQM score is a weighted sum of \emph{error} scores.} with all the scores. This is consistent with the findings above, and also further corroborates the robustness of the collected DA scores.

\subsection{Continuous Rating}
Continuing in this spirit, 
Table~\ref{tab:cr_en_de} compares the correlation of the Continuous Rating (CR) evaluation with the new Direct Assessment evaluations, and the two automatic metrics \textsc{chrF} and COMET on English-to-German simultaneous translations. The correlation between CR and DA of 0.95 demonstrates the validity of the DA scores. The correlation of the automatic metric COMET with DA is 0.94, which is relatively close to the correlation of 0.96 between COMET and DA (see Table~\ref{tab:da_sys_corr_with_comet}). Interestingly, the automatic metric \textsc{chrF} seems to correlate with CR much less than \textsc{chrF} with DA (0.75 vs. 0.93).

To summarize: \textbf{Comparison with both MQM and CR scores verifies the reliability of the DA scores, collected as described in Section~\ref{sec:da}.}
\section{Further Analysis}

\subsection{New versus Old References}
\label{sec:new-vs-old}

As indicated in Section~\ref{sec:ted_data}, TED data included two kinds of references: The original references in subtitle style, and new references that were created for more naturalness. We now turn to the question of whether creating these new references helped make automatic evaluation more reliable. To this end, Table~\ref{tab:da_sys_corr_with_comet} shows correlations for the original references (rows 8--9), addition to the new references (rows 4--5).  We find that correlation coefficients do not differ much between the two references. This indicates that the subtitle-style references are of adequate quality in the context of our data set, and that the effort to create more natural references is perhaps questionable. However, we stress that the situation might change if systems get more accurate, or if there are more systems that are very close in terms of accuracy. In both of these cases having high-quality references is expected to be beneficial for reliable evaluation.

To summarize: \textbf{In the context of our data, the original TED subtitles are equally suitable for evaluation as the new, natural TED references.}

\subsection{Cross-Condition Correlations}
\begin{table*}[tb]
\centering
\begin{tabular}{lcccccccc}
\toprule
        &          &         &                                &                              & \multicolumn{2}{c}{}                 & \multicolumn{2}{c}{\small Original TED refs.} \\
\small Task    & \small Lang.    &\small  Dom.    & $\rho_{\textit{\small DA}/\textsc{\small chrF}}$ & $r_{\textit{\small DA}/\textsc{\small chrF}}$ & $\rho_{\textit{\small DA}/\textsc{\small COMET}}$ & $r_{\textit{\small DA}/\textsc{\small COMET}}$ & $\rho_{\textit{\small DA}/\textsc{\small chrF}}$ & $r_{\textit{\small DA}/\textsc{\small chrF}}$ \\
\midrule

\small Offline & \small en-de    & \small Avg      & $\mathbf{0.98}$ & $\mathbf{0.75}$ & $\mathbf{0.99}$ & $\mathbf{0.89}$ & $\mathbf{0.99}$ & $\mathbf{0.75}$  \\
\small Offline & \small en-ja    & \small Avg      & $\mathbf{0.99}$ & $0.70$          & $\mathbf{1.00}$ & $\mathbf{0.90}$ & $\mathbf{0.99}$ & $0.70$           \\
\small Offline & \small en-zh    & \small Avg      & $\mathbf{0.98}$ & $\mathbf{0.86}$ & $\mathbf{1.00}$ & $\mathbf{1.00}$ & $\mathbf{0.98}$ & $\mathbf{0.96}$  \\
\bottomrule
\end{tabular}
\caption{Correlation of DA and \textsc{chrF} after \textit{averaging} scores from TED and ACL domains.}
\label{tab:da_corr_domain_avg}
\end{table*}

\begin{table*}[tb]
\centering
\begin{tabular}{lcccccccc}
\toprule
         &        &        && & \multicolumn{2}{c}{}               & \multicolumn{2}{c}{\small Original TED references}    \\
Lang. & Dom. & Sys. & $\rho_{\textit{\small DA}/\textsc{\small chrF}}$ & $r_{\textit{\small DA}/\textsc{\small chrF}}$ & $\rho_{\textit{\small DA}/\textsc{\small COMET}}$ & $r_{\textit{\small DA}/\textsc{\small COMET}}$ & $\rho_{\textit{\small DA}/\textsc{\small chrF}}$ & $r_{\textit{\small DA}/\textsc{\small chrF}}$ \\
\midrule
\small en-de &\small  TED & $11$ & $\mathbf{0.97}$ & $\mathbf{0.88}$ & $\mathbf{0.97}$ & $\mathbf{0.85}$  & $\mathbf{0.96}$ & $\mathbf{0.86}$  \\
\small en-de &\small  ACL & $10$ & $\mathbf{0.92}$ & $\mathbf{0.70}$ & $\mathbf{0.98}$ & $\mathbf{0.88}$ & -- & --  \\
\small en-ja &\small  TED &  $9$ & $\mathbf{0.96}$ & $\mathbf{0.75}$ & $\mathbf{0.97}$ &  $0.57$ & $\mathbf{0.96}$ & $\mathbf{0.75}$ \\
\small en-ja &\small  ACL &  $8$ & $\mathbf{0.93}$ &         $0.55$  & $\mathbf{0.99}$ & $\mathbf{0.86}$ & -- & --  \\
\small en-zh &\small  TED &  $9$ & $\mathbf{0.91}$ & $\mathbf{0.83}$ & $\mathbf{0.99}$ & $\mathbf{0.93}$ & $\mathbf{0.91}$ & $\mathbf{0.75}$  \\
\small en-zh &\small  ACL & $10$ & $\mathbf{0.97}$ & $\mathbf{0.98}$ & $\mathbf{0.99}$ & $\mathbf{1.00}$ & -- & --  \\
\bottomrule
\end{tabular}
\caption{Correlations of DA and \textsc{chrF} on original/new references, computed over systems across task types (offline+simultaneous for TED, offline+multilingual for ACL).}
\label{tab:da_sys_corr_task_agg}
\end{table*}

\begin{table*}[tb]
\centering
\begin{tabular}{lcccccccc}
\toprule

     &        &         && \multicolumn{2}{c}{}             & \multicolumn{2}{c}{\small Original TED refs.} \\
Task     &  Language   & $\rho_{\small_{\textit{DA}/\textsc{chrF}}}$ & $r_{\small_{\textit{DA}/\textsc{chrF}}}$ & $\rho_{\small_{\textit{DA}/\textsc{COMET}}}$ & $r_{\small_{\textit{DA}/\textsc{COMET}}}$ & $\rho_{\small_{\textit{DA}/\textsc{chrF}}}$ & $r_{\small_{\textit{DA}/\textsc{chrF}}}$ \\[5pt]
\midrule

\small Offline &\small  en-de  & $\mathbf{0.74}$ & $\mathbf{0.53}$ & $-0.30$ & $-0.01$ & $\mathbf{0.66}$ & $0.31$ \\
\small Offline &\small  en-ja  & $\mathbf{0.76}$ & $\mathbf{0.64}$ & $~~~0.00$ & $~~~0.39$ & $\mathbf{0.84}$ & $\mathbf{0.64}$ \\
\small Offline &\small  en-zh  & $\mathbf{0.90}$ & $\mathbf{0.74}$ & $-0.36$ & $-0.03$ & $0.34$ & $0.20$ \\

\bottomrule
\end{tabular}
\caption{Correlation of DA scores and automatic metrics, computed over systems across both data domains. We show only the offline task, the only task that includes both domains.}
\label{tab:da_sys_corr_data_agg}
\end{table*}

\begin{table*}[tb]
\centering
\begin{tabular}{lcccccccc}
\toprule
     &        &         && \multicolumn{2}{c}{}             & \multicolumn{2}{c}{\small Original TED refs.} \\
Task & Domain & $\rho_{\small_{\textit{DA}/\textsc{chrF}}}$ & $r_{\small_{\textit{DA}/\textsc{chrF}}}$ & $\rho_{\small_{\textit{DA}/\textsc{COMET}}}$ & $r_{\small_{\textit{DA}/\textsc{COMET}}}$ & $\rho_{\small_{\textit{DA}/\textsc{chrF}}}$ & $r_{\small_{\textit{DA}/\textsc{chrF}}}$ \\[5pt]
\midrule
\small Offline & \small TED  & $\mathbf{0.87}$ & $\mathbf{0.94}$ & $\mathbf{0.70}$ & $\mathbf{0.52}$ & $\mathbf{0.67}$ & $\mathbf{0.67}$ \\
\small Multi & \small ACL  & $\mathbf{0.67}$ & $0.60$ & $0.58$ & $0.49$ & -- & -- \\
\small Offline & \small ACL  & $\mathbf{0.78}$ & $\mathbf{0.78}$ & $\mathbf{0.65}$ & $\mathbf{0.45}$ & -- & -- \\
\small Simul & \small TED  & $\mathbf{0.88}$ & $\mathbf{0.93}$ & $\mathbf{0.69}$ & $0.01$ & $\mathbf{0.79}$ & $\mathbf{0.82}$ \\

\bottomrule
\end{tabular}
\caption{Correlation of DA scores and \textsc{chrF}, computed over systems across all three language pairs.}
\label{tab:da_sys_corr_lang_agg}
\end{table*}

The overlap between conditions such as task, domain, and languages in our data provides an opportunity to study the feasibility of cross-condition comparisons. 

First, we combine \textit{domains} by using the average \textsc{chrF} score between ACL and TED test sets for the offline task. Such a comparison could be conducted to rank systems with regards to their cross-domain translation ability. It also provides an additional perspective to verify observations made in earlier sections. To this end, we observe in Table~\ref{tab:da_corr_domain_avg} that differences between original and new TED references exist but are minor, confirming the conclusions from Section~\ref{sec:new-vs-old}. The Table also confirms COMET to be better correlated with human judgments than \textsc{chrF}, similar to findings in Section~\ref{sec:results-da-automatic}.

Another question we might pose is whether automatic metrics can be compared across different conditions. In speech translation, especially the case of comparing across task types (offline, multilingual, simultaneous) is of practical relevance, e.g.\ where one might wish to judge how well a given offline system is perceived by users in comparison to a given simultaneous system. Practitioners might also wish to compare systems across domains or even languages, for example if through experience a threshold on an automatic metric has been determined above which user experience is satisfactory, and one wonders whether the same threshold can be applied to a different domain or target language. Factors that would hinder such a comparison include any phenomenon that systematically biases an automatic score in comparison to human judgement, such as morphological complexity (when comparing across target languages), translation artifacts introduced by simultaneous translation systems (when comparing task types), or peculiarities of the domain that have an overly strong or weak influence on the metric. While researchers often refrain from such comparisons due to potential issues along these lines, our data, due to its overlapping nature, provides some opportunity to investigate the validity of such comparisons. 

We conduct this analysis by simply computing the correlation between DA and automatic metric jointly across systems from the different conditions. 
First, we observe that jointly considering systems from different \textit{task conditions} (offline, multilingual, simultaneous) yields high correlations (Table~\ref{tab:da_sys_corr_task_agg}). As discussed, this is desirable in speech translation with its multitude of task types. 

Next, doing the same analysis across domains (Table~\ref{tab:da_sys_corr_data_agg}) and target languages (Table~\ref{tab:da_sys_corr_lang_agg}) reveals some interesting findings: Correlations in rows 3--4 (\textsc{chrF} with new TED references) are significantly lower in comparison but still strongly positive, indicating that these comparisons, with some care, can be informative. In contrast, \textsc{chrF} against subtitle references, and COMET, are more poorly correlated. This indicates that more accurate TED references are of value under such challenging conditions, and that COMET is more brittle than \textsc{chrF}.

To summarize: \textbf{We find further evidence for observations from previous sections, and that it is safe to compare systems across the common speech translation task types such as offline and simultaneous systems.}

\section{Conclusion}
We have presented a first step towards establishing human evaluation practices in speech translation. Our approach includes direct assessment based on automatically segmented inputs. Comparison with MQM and CR shows the robustness of our approach. Correlations with automatic metrics are generally high, and (in line with findings in text translation) COMET is observed to slightly outperform \textsc{chrF}, but more experiments are needed to corroborate these findings. 

Future work may investigate whether evaluation in which human annotators are presented with audio segments rather than text segments could be a feasible improvement.

\nocite{*}
\section{Bibliographical References}
\label{sec:reference}

\bibliographystyle{lrec-coling2024-natbib}
\bibliography{custom} 

\pagebreak
\clearpage

\appendix

\section{Appendix: Annotator instructions}

in order to minimize the impact of segmentation issues during direct assessment (DA), the following instructions were given to human annotators: \\

\textit{Sentence boundary errors are expected and should not be factored in when judging translation quality. This is when the translation appears to be missing or adding extra words but the source was segmented at a different place. To this end, we have included the translations for the previous and 
next sentences also. If the source and translation are only different because of sentence boundary
 issues, do not let this affect your scoring judgment.}

\end{document}